\documentclass[letterpaper, 10 pt, conference]{ieeeconf}  

\IEEEoverridecommandlockouts                              

\usepackage{graphics} 
\usepackage{epsfig} 

\usepackage{mathtools}
\usepackage{amsmath} 
\usepackage{amssymb}  
\usepackage{todonotes}
\usepackage{graphicx}
\usepackage{nicematrix}
\usepackage{caption,subcaption}
\usepackage{mathrsfs}
\usepackage{ctable} 
\usepackage{url}

\usepackage{multirow}
\usepackage{algorithm}
\usepackage{algpseudocode}
\usepackage[normalem]{ulem}
\usepackage[dvipsnames]{xcolor}


\begin{document}
\title{HERE: Hierarchical Active Exploration of Radiance Field with Epistemic Uncertainty Minimization}
\author{
Taekbeom Lee$^{*}$,
    \thanks{ $^{*}:$ Equal contribution. The authors are with the Department of Aerospace Engineering, Seoul National University, Seoul 08826, South Korea (e-mail: {ltb1128, dabin404, duscjs59, hjinkim}@snu.ac.kr, corresponding author: H. Jin Kim). 
    This work was supported by Samsung Research Funding \& Incubation Center of Samsung Electronics under Project Number SRFC-IT2402-17.}
Dabin Kim$^{*}$, Youngseok Jang, and H. Jin Kim
}

\maketitle

\begin{abstract}
We present \textit{HERE}, an active 3D scene reconstruction framework based on neural radiance fields, enabling high-fidelity implicit mapping.
Our approach centers around an active learning strategy for camera trajectory generation, driven by accurate identification of unseen regions, which supports efficient data acquisition and precise scene reconstruction.
The key to our approach is epistemic uncertainty quantification based on evidential deep learning, which directly captures data insufficiency and exhibits a strong correlation with reconstruction errors. 
This allows our framework to more reliably identify unexplored or poorly reconstructed regions compared to existing methods, leading to more informed and targeted exploration.
Additionally, we design a hierarchical exploration strategy that leverages learned epistemic uncertainty, where local planning extracts target viewpoints from high-uncertainty voxels based on visibility for trajectory generation, and global planning uses uncertainty to guide large-scale coverage for efficient and comprehensive reconstruction.
The effectiveness of the proposed method in active 3D reconstruction is demonstrated by achieving higher reconstruction completeness compared to previous approaches on photorealistic simulated scenes across varying scales, while a hardware demonstration further validates its real-world applicability.
\end{abstract}    
\section{Introduction}
\label{sec:intro}

Accurately and efficiently representing 3D environments is a fundamental challenge in robotics, computer vision, and graphics. 
While traditional methods such as point clouds and meshes are widely used, radiance field-based approaches such as Neural Radiance Fields (NeRF) \cite{mildenhall2021nerf} and 3D Gaussian Splatting (3DGS) \cite{kerbl20233d} offer superior detail and lighting capture.
Most research focuses on reconstructing scenes from given datasets, yet achieving high-quality results that support navigation, interaction, and downstream perception tasks demands both comprehensive scene coverage and detailed observation of fine-grained structures.

Active 3D reconstruction addresses this challenge by directing robot motion to acquire high-quality data.
Uncertainty can serve as an effective cue to identify under-reconstructed or unexplored regions that require further observations, while planning provides the means to ensure that these regions are actually and efficiently observed.
The objective of this letter is to propose an integrated system that enhances scene coverage, reduces computational overhead, and achieves high reconstruction accuracy.
Despite the growing interest in 3DGS methods \cite{jin2025activegs, li2025activesplat}, we adopt NeRF for active reconstruction because NeRF yields smooth surfaces and usable signed distance fields for downstream tasks, whereas obtaining these reliably from 3DGS remains nontrivial. Additionally, it is more memory‑efficient, which is especially important for robots with limited resources.

\begin{figure}[t!]
  \centering
  \includegraphics[width=0.9\columnwidth]{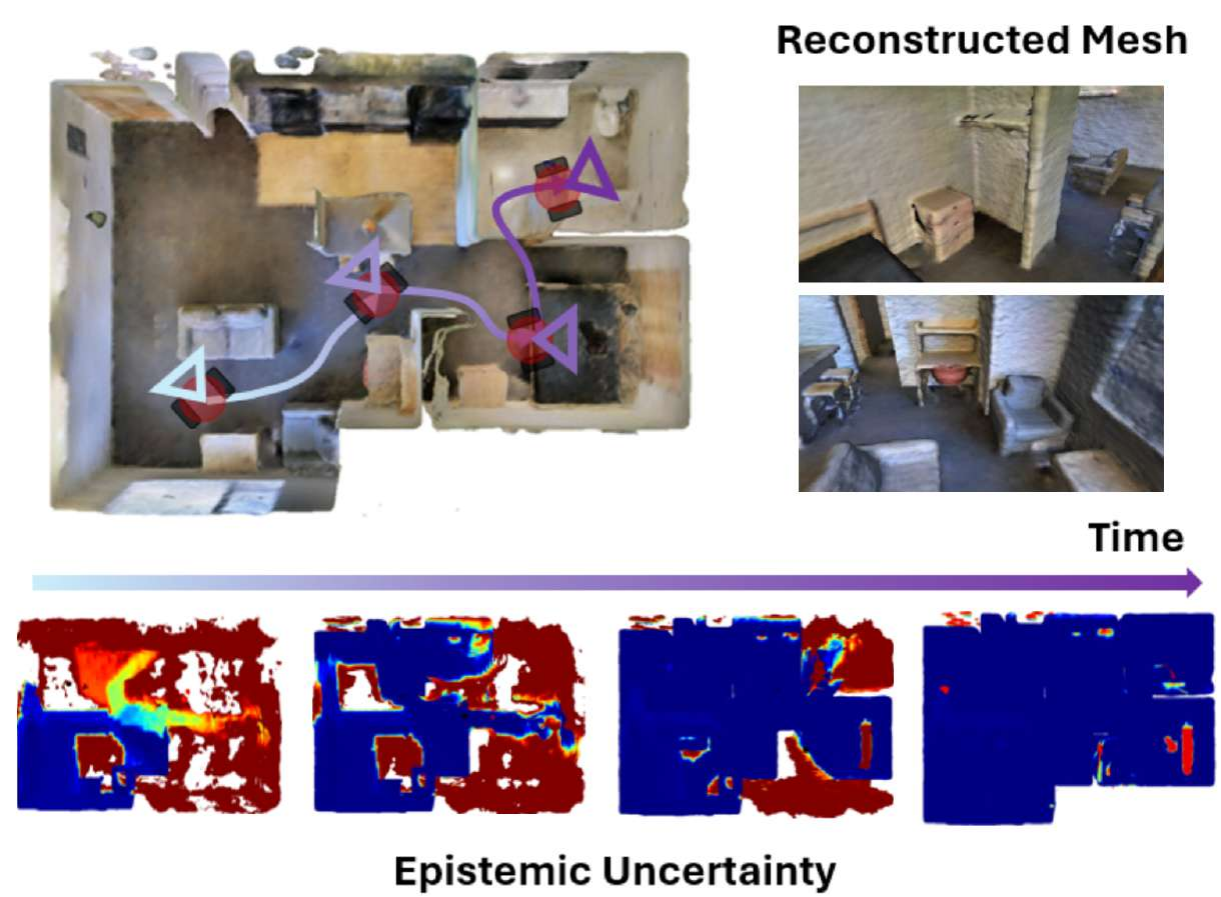}
  \caption{
  Our approach performs active scene reconstruction of neural radiance fields by leveraging epistemic uncertainty to guide exploration of an agent.
  It identifies regions with high epistemic uncertainty and generates an informative path. This, combined with hierarchical planning, ensures both scene coverage and detailed exploration of uncertain areas, enabling high-quality reconstruction across various scales.
  }
  \vspace{-6mm}
  \label{fig:teaser}
\end{figure}

According to the extensive literature in active SLAM \cite{placed2023survey} and active learning \cite{ren2021survey}, uncertainty estimation plays a pivotal role in guiding exploration and improving reconstruction efficiency. 
Existing approaches \cite{feng2024naruto, ran2023neurar, lee2022uncertainty} integrate Bayesian modeling into NeRF to estimate uncertainty and optimize training view selection. However, they focus on overall uncertainty without explicitly differentiating epistemic uncertainty, which reflects the model’s knowledge limitations, from aleatoric uncertainty, which stems from irreducible data noise. Approaches that quantify epistemic uncertainty, such as network parameter perturbation \cite{yan2023active} and Fisher information analysis \cite{jiang2024fisherrf, xu2024hgs}, are impractical to use in real-time planning due to expensive computation.

In the second aspect of active reconstruction, we need to design an informative path planning strategy that fully covers the scene without leaving unmapped areas while selecting informative views. 
Recent active learning approaches for NeRF \cite{yan2023active, feng2024naruto} employ sampling‑based goal search to cope with the high dimensionality of planning, but this often fails to ensure full coverage. Moreover, focusing on information gain solely at the goal states \cite{yan2023active, feng2024naruto, kuang2024active} and ignoring the information accumulated along the continuous camera trajectory can lead to suboptimal, uninformative behavior.

In this work, we propose \textit{HERE}, a system that integrates a neural mapping module with epistemic uncertainty-driven informative path planning. 
We develop a novel uncertainty quantification (UQ) method for radiance fields that enables the use of evidential deep learning (EDL) to accurately assess how well a region has been observed, thereby allowing viewpoint evaluation based on informativeness.
We demonstrate through experiments that our uncertainty estimates exhibit stronger correlation with ground truth coverage compared to representative active mapping methods.
By storing spatial uncertainty in learnable grids, our system ensures real-time performance because estimation requires only trilinear interpolation, while ensuring stable long-term learning as updates affect only nearby voxels.
Furthermore, inspired by batch active learning \cite{kirsch2019batchbald}, our method considers joint information along the camera trajectory rather than combining separately computed information from individual views, leading to a more accurate approximation of the total information gathered.
To ensure both high reconstruction accuracy and comprehensive scene coverage, we introduce a hierarchical planning framework. 
This structure balances global exploration with local planning, making our method effective across both small- and large-scale environments.
In summary, our key contributions are as follows:
\begin{itemize}
    \item An integrated real-time active mapping framework that guides reconstruction using uncertainty-aware planning.
    \item A novel approach to quantify epistemic uncertainty in neural implicit mapping by extending evidential deep learning to radiance fields.
    \item A planner that ensures global coverage while refining local trajectories via epistemic uncertainty for scalable reconstruction.
    \item State-of-the-art reconstruction completeness on datasets of various scales, with a hardware demonstration validating real-world applicability.
\end{itemize}


\section{Related Work}
\label{sec:relwork}

\subsection{Active Scene Reconstruction}
Active planning for efficient scene reconstruction has been widely studied in robotics and computer vision \cite{placed2023survey}. While many prior methods rely on explicit representations such as meshes or occupancy grids \cite{carrillo2015autonomous, tabib2021autonomous}, the emergence of NeRFs as a powerful tool for high-fidelity 3D reconstruction has shifted recent efforts toward developing active learning strategies for radiance field mapping.

Several studies \cite{lee2022uncertainty, pan2022activenerf, ran2023neurar} integrate NeRF with Next-Best-View (NBV) strategies by sampling candidate views and computing information gain via rendering. However, the high computational cost of NeRF rendering hinders real-time performance.
\cite{feng2024naruto} partially reduces the computational cost by adopting hybrid scene representations \cite{wang2023coslam}.
Frame-level view selection is simplified into goal position search in \cite{yan2023active, feng2024naruto}. 
However, these methods restrict the action space to limited domains such as a hemisphere or a 2D plane, or neglect information gain throughout the trajectory.
Unlike prior approaches, our method efficiently computes viewpoint information gain with explicit consideration of orientation and extends beyond frame-level selection to generate informative trajectories.
\vspace{-1mm}

\begin{figure*}[t]
  \centering
  \includegraphics[width=1.7\columnwidth]{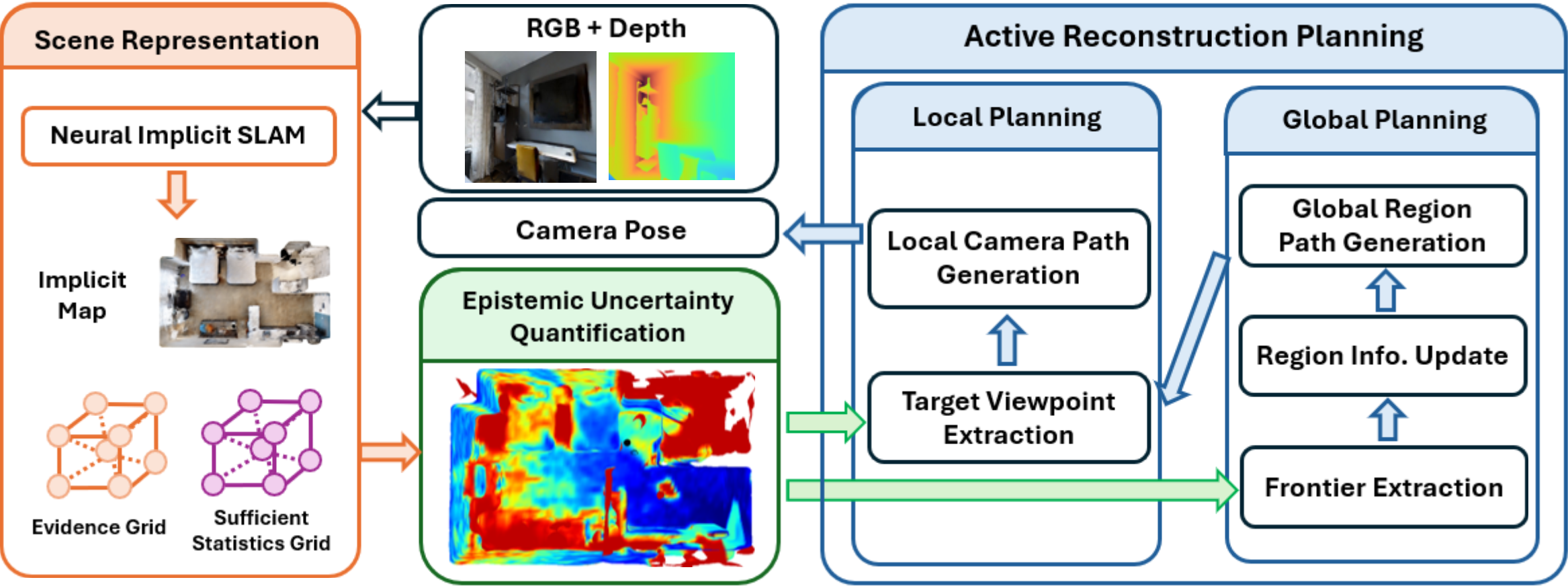}
  \caption{An overview of the framework. Given RGB and depth images, the neural implicit SLAM module learns an implicit map along with evidence and sufficient statistics grids, which are used to quantify epistemic uncertainty. The hierarchical active reconstruction planning method leverages this uncertainty for target view selection in local planning and frontier extraction for global region path generation. The planner then generates a camera trajectory to explore the unseen environment. }
  \vspace{-3mm}
  \label{fig:planning_overview}
\end{figure*}

\subsection{Uncertainty in Radiance Field}
Regarding UQ in NeRF, several works utilize NeRF-specific proxies such as rendering weight \cite{lee2022uncertainty}.
Techniques from traditional deep learning, such as modeling NeRF predictions as Gaussian, are adapted \cite{ran2023neurar, pan2022activenerf}.
In image-based neural rendering, \cite{jin2023neu} quantifies uncertainty inherited from input for next-best-view planning to acquire informative reference views.
However, their UQ includes aleatoric uncertainty, arising from factors such as transient objects or changes in lighting conditions, which are irreducible by collecting additional observations.

Epistemic uncertainty, caused by insufficient data, serves as a meaningful signal for active reconstruction in NeRF.
Several studies try to quantify such uncertainty, particularly in the presence of missing views or occluded regions.
Stochastic NeRF \cite{shen2022conditional} quantifies uncertainty via Monte Carlo approximation, while \cite{sunderhauf2023density} uses an ensemble model, but multiple predictions require large computation. 
Recent works \cite{goli2024bayes, jiang2024fisherrf} estimate epistemic uncertainty using the Laplace approximation.
However, these methods are not suitable for real-time active reconstruction due to the high computational cost of back-propagation. Moreover, they rely on approximate posterior distributions over model parameters, which can limit the accuracy of uncertainty estimation.
EDL provides an alternative framework for modeling epistemic uncertainty by predicting a prior over the predictive distribution. 
\cite{chen2024nerf_localization} showed effectiveness of EDL to model reliability of 2D-3D correspondences. For active reconstruction, we propose a novel approach that estimates the uncertainty of the geometry modelled by the radiance field. We update the prior over 3D geometry for effective identification of undertrained regions, and use explicit learnable grids for fast epistemic uncertainty estimation via trilinear interpolation.

\vspace{-1mm}

\section{Preliminaries}
\label{sec:prem}
\subsection{Evidential Deep Learning for Regression}
To quantify uncertainty in regression, predictions are often modeled with a Gaussian distribution.
However, a high variance alone cannot distinguish whether the data is noisy or the model is uncertain.
Evidential Deep Learning (EDL) addresses this by placing a prior over the Gaussian parameters $\theta\!=\!(\mu, \sigma^2)$ using a Normal Inverse-Gamma (NIG) distribution with parameters $m\!=\!(\mu_0,\lambda,\alpha,\beta)$, in order, mean, precision, shape parameter, and scale parameter.
Since NIG is conjugate to the Gaussian, the posterior given observations $\mathbb{D}\!=\!\{ y_i\}_{i=1}^{N}$ remains NIG, as described by the Bayesian update rule:
\begin{align}
    \label{eq:nig}
    p(\theta \mid \mathbb{D}, m^{post}) &\propto \left[\prod_{i=1}^{N}\mathcal{N}(y_i \mid \theta)\right]\mathrm{NIG}(\theta \mid m^{pri}) \notag\\
    &= \mathrm{NIG}(\theta \mid m^{post}),
\end{align}
where the posterior parameters $m^{post}$ can be obtained in closed form \cite{murphy2007conjugate}.
Eq. \eqref{eq:nig} can be simplified by defining $\chi=(\mu_0, \mu_0^2 + \beta/\alpha)$ and setting $n=\lambda=2\alpha$ \cite{charpentier2021natural}, leading to 
\begin{align} \label{eq:bayesian}
    \chi^{post} = \frac{n^{pri}\chi^{pri} + N\chi^{\mathbb{D}}}{n^{pri} + N},\quad n^{post}=n^{pri}+N,
\end{align}
with $\chi^{\mathbb{D}}=\left(\bar{y}, \frac{1}{N}\sum_i y_i^2\right)$. 
$\chi^{\mathbb{D}}$, and $\chi^{pri}$, $\chi^{post}$ are sufficient statistics for estimating $\theta$ from the observations $\mathbb{D}$, pseudo-observations of the prior and posterior data, respectively.
$n^{pri}$ and $n^{post}$ are the numbers of pseudo-observations, also referred to as evidence.
The updated posterior parameters $m^{post}$ allow epistemic uncertainty to be predicted in closed form, as detailed in Sec. \ref{sec:uq}.

\subsection{Neural implicit mapping}
\label{sec:map representation}
Co-SLAM \cite{wang2023coslam} represents 3D geometry by predicting colors and truncated signed distance fields (TSDF) for world coordinates.
It uses both coordinate encoding \cite{muller2019neural} and sparse parametric encoding \cite{mueller2022instant} along with a geometry decoder and color MLP, enabling fast, high-fidelity reconstruction.
Bundle adjustments are executed by minimizing several objectives for sampled rays. 
Along each ray $r$ defined by camera origin $o$ and ray direction $d$, 3D points are sampled at predefined depths $t_i$ as $x_i=o+t_id$.
Rendering weights $w_i$ are computed from predicted TSDFs $s_i$, and color and depth are rendered by a weighted sum of predicted colors $c_i$ and $t_i$.
Rendering losses compare the rendered color and depth with the observed color and depth.
To achieve accurate and smooth reconstruction, SDF losses $\mathcal{L}_{sdf}$ and $\mathcal{L}_s$ are applied to points inside ($\left | D_r-t_i \right | \leq tr$) and outside the truncation region, respectively.
\begin{align}
    \mathcal{L}_{sdf} &= \eta \sum_{r \in R_d} \eta_{r} 
    \sum_{i \in S_r^{tr}}(s_i - (D_r-t_i))^2, \label{eq:sdf loss} \\
    \mathcal{L}_{fs} &= \eta \sum_{r \in R_d} \eta_{r} 
    \sum_{i \in S_r^{fs}}(s_i - tr)^2, \label{eq:free space loss}
\end{align}
where $D_r$ is observed depth for the ray $r$, $S_r^{(.)}$ is the set of sampled points in each region, $tr$ is the truncation distance, and $\eta_{r}=\frac{1}{|S_r^{tr}|}$. 
Feature smoothness loss is performed for smooth reconstruction in unobserved free spaces.
For further details, we refer readers to Co-SLAM.
\section{Epistemic Uncertainty Quantification}
\label{sec:uq}
To effectively capture epistemic uncertainty, we integrate EDL-based uncertainty quantification into neural implicit mapping.
Prior works on NeRF uncertainty often model 3D color as Gaussian, $c_i \sim \mathcal{N}(\bar{c}_i,\sigma_i^2)$, and train the model through the rendered color which also follows a Gaussian $\hat{C}=\sum_{i=1}^Mw_ic_i \sim \mathcal{N}(\sum_{i=1}^Mw_i\bar{c}_i,\sum_{i=1}^Mw_i^2\sigma_i^2).$
However, this strategy does not extend to evidential models: when a NIG prior is placed on each $c_i$, the resulting distribution of $\hat{C}$ no longer has a tractable form. 
This breaks the conjugacy required for tractable posterior updates, which is an assumption essential for EDL.

\begin{figure}[t!]
        \centering
		\includegraphics[width=0.95\columnwidth]{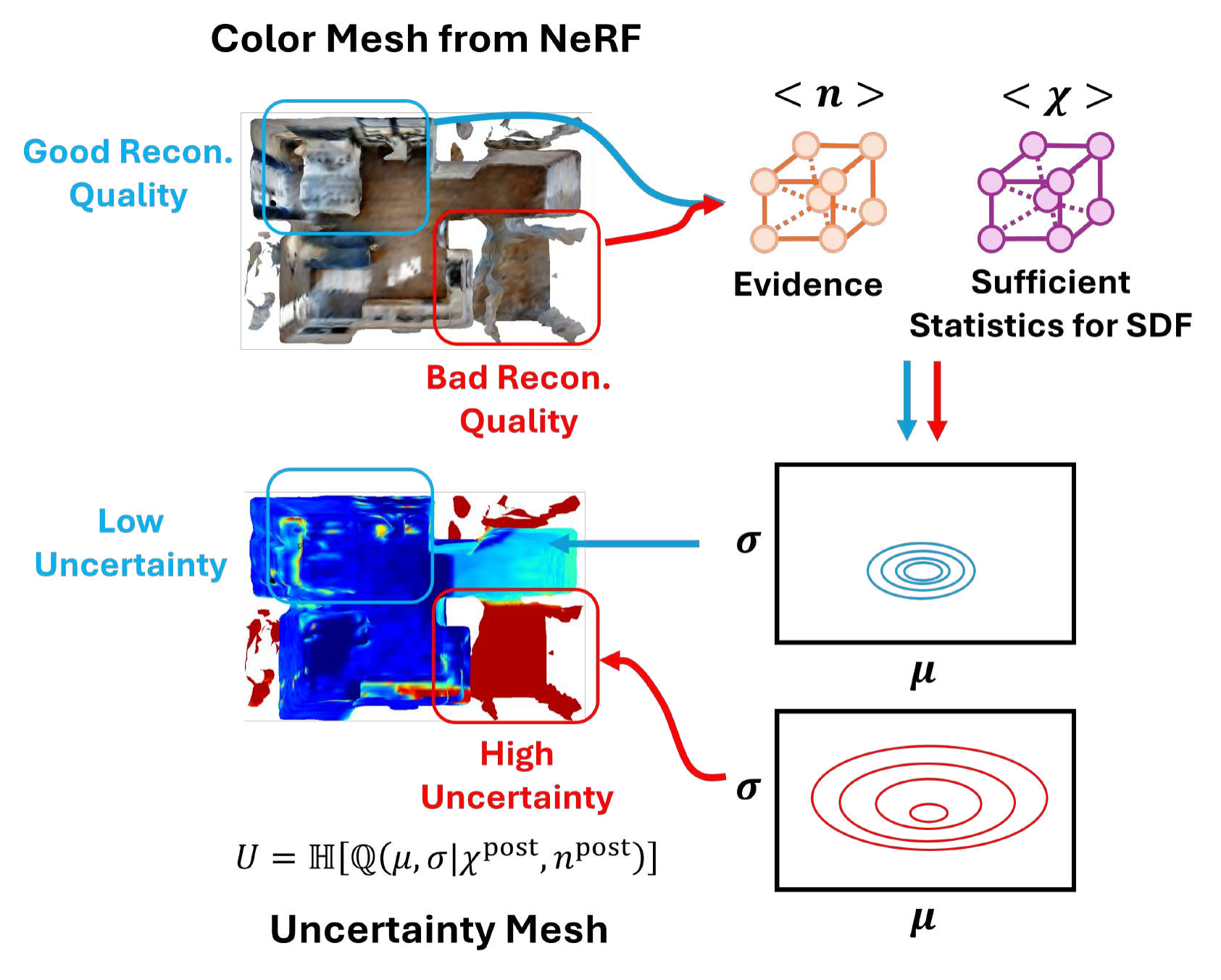}
		\caption{ Illustration of
        our uncertainty quantification module. 
        The spread around $(\mu,\sigma)$ enlarges in poorly reconstructed regions, which we model as epistemic uncertainty.}
        \vspace{-5mm}
		\label{fig:uq_explain}
\end{figure}

Alternatively, we extend Co-SLAM’s SDF prediction to probabilistic modeling.
We model the predictive distribution as a Gaussian $\mathcal{N}(s \mid \mu, \sigma^2)$, with a NIG prior defined over its parameters $\mu$, $\sigma^2$.
Our model predicts the posterior update $\chi_{i},n_{i}$ for each 3D point $x_i$, effectively replacing the observation-based Bayesian update in eq. \eqref{eq:bayesian}.
\begin{equation}\label{eq:posterior update}
    \chi^{post}_i = \frac{n^{pri}\chi^{pri} + n_{i}\chi_{i}}{n^{pri} + n_{i}}, \hspace{5mm} n^{post}_i = n^{pri} + n_{i}.
\end{equation}
where we set $\chi^{pri}\!=\!(0, 3)$, corresponding to a prior SDF with zero mean and variance 3, with $n^{pri}\!=\!1$. 
We treat $s_i$ predicted by the original Co-SLAM as the first statistic $\chi_{i,1}$, while a separate uncertainty learning module predicts the second statistic $\chi_{i,2}$ and the evidence $n_{i}$.
This enables uncertainty modeling without modifying the neural implicit mapping, preserving reconstruction quality and enabling modular integration with various mapping methods.

We implement the uncertainty learning module using two learnable grids, $V_{\rho}$ and $V_{\tau}$, from which 3D point $x_i$ queries a confidence score $\rho$ and second moment $\tau\!=\!\frac{\beta}{\alpha}$ via trilinear interpolation. 
We compute $n_{i}=N_S*\varphi(V_{\rho}(x_i))$ and $\tau_{i}=V_{\tau}(x_i)$, where $\varphi$ is the sigmoid function and $N_S$ is a scale constant.
From $s_i$, $n_{i}$, and $\tau_{i}$, we obtain posterior NIG parameters $m^{post}_i\!=\!(s_i,\lambda^{post}_i,\alpha^{post}_i,\beta^{post}_i)$ via eq. \eqref{eq:posterior update} and the definition of $n$ and $\chi$ in Sec. \ref{sec:prem}.
The overview of quantifying epistemic uncertainty is described in Fig. \ref{fig:uq_explain}.
Our grid-based representation ensures spatial locality, restricting uncertainty learning to local regions.
This is especially beneficial in incremental mapping, where new observations affect only nearby voxels, preserving past estimates and ensuring stable long-term learning.

We employ the entropy of NIG distribution to quantify epistemic uncertainty of each 3D point.
It reflects how much the posterior NIG is uncertain in estimating the predictive Gaussian, which aligns with uncertainty caused by insufficient information about the data. 
\begin{align} \label{eq:epistemic}
    &u_{epi}(\lambda,\alpha,\beta)=\mathbb{H}[\mathrm{NIG}(\theta \mid m)] \notag \\
    &= \log\left(\frac{(2\pi)^{\frac{1}{2}}\beta^{\frac{3}{2}}\Gamma(\alpha)}{\lambda^{\frac{1}{2}}}\right) - \left(\alpha+\frac{3}{2}\right)\psi(\alpha) + \alpha + \frac{1}{2}
\end{align}

Our uncertainty learning module is jointly trained with neural implicit mapping during bundle adjustment.
We extend the SDF losses (eq. \eqref{eq:sdf loss} and \eqref{eq:free space loss}) to a Bayesian formulation \cite{charpentier2021natural}, encouraging the predicted posterior to align with the true posterior.
For each 3D point with ground truth SDF $s_{gt}$, the Bayesian loss is derived from the NIG properties:

{\small
\begin{align} \label{eq:bayesian loss}
    &\mathcal{L}^*\left(s_{gt},m\right) = \mathbb{E}_{\theta \sim \mathrm{NIG}(\theta \mid m)}[-\log \mathcal{N}(s_{gt} \mid \theta)] - \gamma \mathbb{H}[\mathrm{NIG}(\theta \mid m)] \notag \\
    &= \frac{1}{2}\left( \frac{\alpha}{\beta} (s_{gt} - s)^2 + \frac{1}{\lambda} - \psi(\alpha) + \log{2\pi\beta} \right) - \gamma u_{epi}(\lambda,\alpha,\beta)
\end{align}
}
The ``post" superscript is omitted for clarity, but the loss and the epistemic uncertainty are computed with the posterior parameters $m^{post}$.
The entropy term (eq. \eqref{eq:epistemic}) also serves as a regularizer, penalizing overconfident posteriors.
We set $s_{gt}$ to $D_r-t_i$ and $tr$ for points inside and outside the truncation region, respectively, leading to the Bayesian SDF losses:
\begin{align}
    \mathcal{L}_{sdf}^* &= \eta \sum_{r \in R_d} \eta_{r}  \sum_{i \in S_r^{tr}}\mathcal{L}^* (D_r-t_i, m^{post}) \\
    \mathcal{L}_{sdf}^* &= \eta \sum_{r \in R_d} \eta_{r}  \sum_{i \in S_r^{tr}}\mathcal{L}^* (tr, m^{post})
\end{align}
These losses are applied with other Co-SLAM losses.
\vspace{-2pt}

\section{Active Scene Reconstruction}
\label{sec:planning}
\vspace{-1pt}



An effective active learning strategy for scene reconstruction must satisfy two key requirements: (1) selecting views that target areas of high uncertainty, and (2) achieving complete scene coverage. 
To address the first requirement, our local planner (LP) generates viewpoints focusing on high-uncertainty regions. However, relying solely on the LP may limit exploration to previously visited areas. To overcome this, we introduce a global planner (GP) that generates paths covering discrete regions, ensuring comprehensive exploration.

\subsection{Scene Coverage Planning}
\subsubsection{Environment Decomposition}
Global scene planning is computationally demanding, and prior sampling-based methods  \cite{feng2024naruto,yan2023active} often yield incomplete coverage. To address this, we propose a coverage-based GP that partitions the environment into regions, $S_{R}=\{R_{i}\}_{i=1}^{N}$, where each representative point (RP) is the centroid of unoccupied voxels. The environment is first uniformly partitioned into fixed-resolution regions within the specified 3D environment bounds.
A connectivity graph $\mathcal{G}_{R}=(S_{R},E_{R})$, inspired by \cite{zhang2024falcon}, encodes  traversability, where edges $E_{R}$ represent connections validated by A$^{*}$ algorithm between their RPs. To prevent losing potential connectivity, high-uncertainty voxels are treated as unoccupied. Regions are classified as \textit{unexplored}, \textit{exploring}, or \textit{explored}. An \textit{unexplored} region contains high-uncertainty voxels based on our UQ; an \textit{exploring} region includes frontier voxels; and an  \textit{explored} region lacks such frontiers. Frontier voxels are identifiedbased on the occupancy of voxels, evaluated through the SDFs, and the uncertainty of neighboring voxels.

\subsubsection{Region-Level Coverage Planning}
For coverage planning, we propose a method to determine the optimal visitation order for \textit{exploring} regions by formulating the problem as an open-loop TSP. The cost function for each edge $e_{ij}$ connecting regions $R_{i},R_{j}$ is defined as $c(e_{ij})=l_{ij}-k\cdot f_{ij}$, where $l_{ij}$ is the distance between  representative points, $f_{ij}$ is the number of frontier voxels in the regions, and $k$ is a weighting parameter. 
Since the problem is NP-complete, we employ the Lin-Kernighan heuristic \cite{lin1973effective} to approximate the solution. The resulting path achieves complete coverage by prioritizing frontier-rich regions and minimizing path length.

\vspace{-1pt}

\subsection{Camera Viewpoint Trajectory Generation}

\begin{figure}[t!]
        \centering
		\includegraphics[width=1.0\columnwidth]{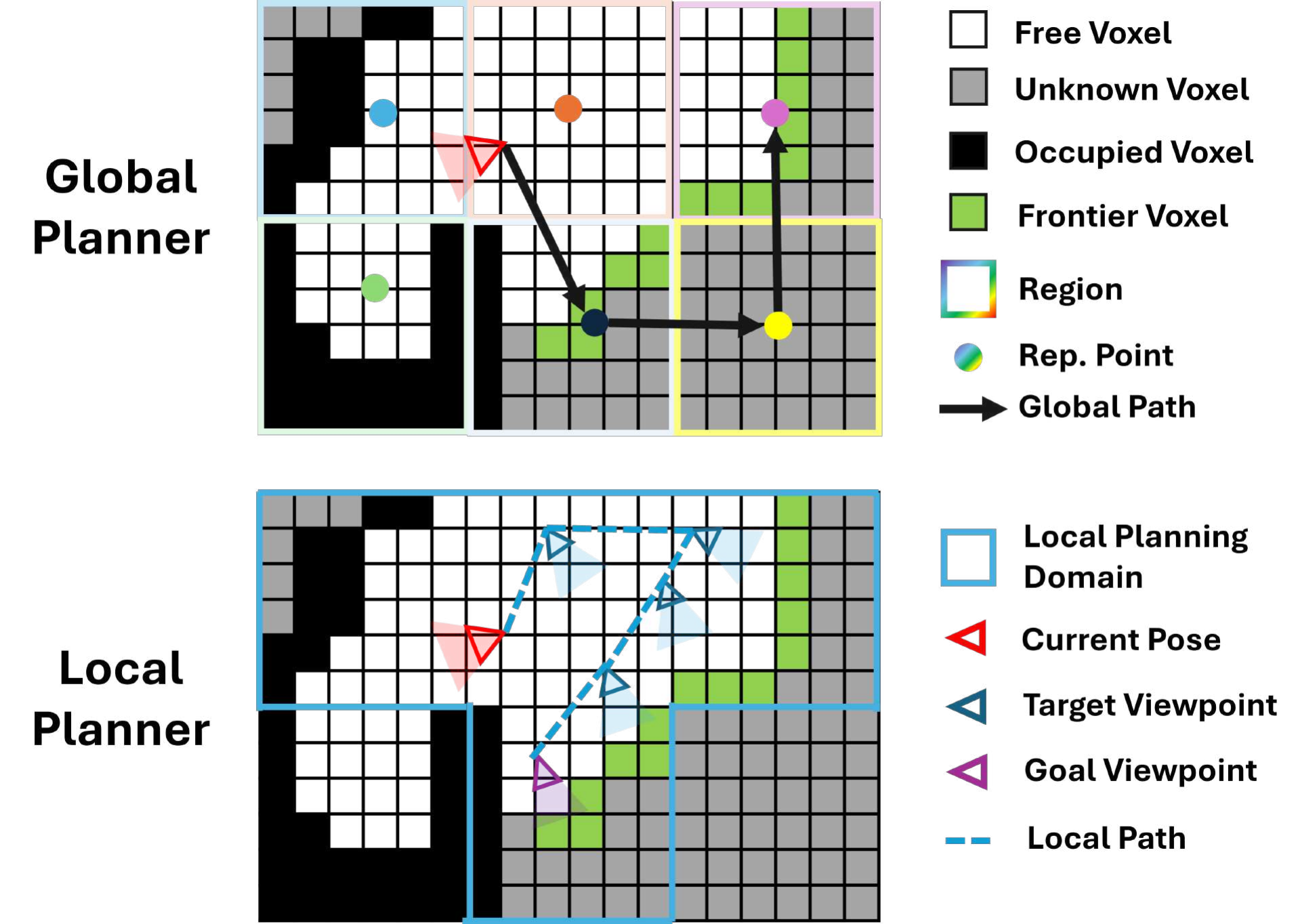}
		\caption{ Illustration of the process of the scene coverage planning and local trajectory planning. 
        }
        \vspace{-5mm}
        \label{fig:planning_explain}
\end{figure}

This subsection details the design of the LP, which refines the global path into an information-rich 6DoF trajectory by selecting and connecting informative camera viewpoints.

\subsubsection{Target Viewpoints Extraction}

To generate an information-rich path, we extract 6DoF target viewpoints with visibility to voxels exhibiting high epistemic uncertainty. The viewpoint space is defined by combining discrete positions in the local planning domain with sampled orientations. Existing approaches such as occupancy-grid raycasting \cite{carrillo2015autonomous} or direct NeRF rendering are computationally expensive and unsuitable for real-time planning. In contrast, our method integrates a NeRF-based representation with a tailored uncertainty accumulation scheme, designed in accordance with our uncertainty quantification framework, thereby enabling efficient real-time viewpoint selection.

The global goal is defined as the representative point of the next region on the global path, and the local planning domain includes the current region and its neighbors. 
Candidate viewpoints are generated by pairing positions sampled regularly within this domain with orientations sampled via Fibonacci sphere sampling \cite{swinbank2006fibonacci}.
For each sampled pose, uncertainty is accumulated as the sum of uncertainties from the top-$k$ high-uncertainty voxels that satisfy: (1) distance constraints, (2) visibility via line-of-sight checks in the SDF, and (3) inclusion within a conic field of view approximated by the camera's mean FoV.

We select viewpoints to maximize the total accumulated uncertainty of the set over chosen viewpoints. This set-based criterion is optimized using a greedy strategy, which is supported by submodular maximization theory \cite{roberts2017submodular, kirsch2019batchbald}.
Viewpoints are iteratively selected when their accumulated uncertainty surpasses a threshold, and the selection process continues until a maximum number of viewpoints is obtained.
To enhance coverage and reduce redundancy, previously selected voxels are filtered out. 
For each selected viewpoint, additional rotation samples at the same position are examined to enhance coverage while minimizing unnecessary movement.

\subsubsection{Camera Viewpoint Trajectory Generation}
We construct a target-viewpoint graph $\mathcal{G}_{V}=(S_{V}, E_{V})$, where $S_{V}$ denotes informative viewpoints and $E_{V}$ contains collision-free edges between them. To guide the camera toward the global goal, we select a local goal as follows: if the global goal lies within the local planning domain, it is used directly; otherwise, 
we select the most informative target viewpoints from the region with the smallest center-to-goal distance.
A TSP is then solved over $\mathcal{G}_{V}$, with edge costs defined by traversal time.  To enforce fixed start and goal, we introduce a dummy node $V_{d}$ connected only to the start and goal with zero cost, ensuring a valid start-to-goal solution.
The ordered target viewpoints are interpolated in position and orientation, resulting in a smooth and continuous camera trajectory.

\subsubsection{Escaping Local Optima}
Although the LP biases the camera toward the global goal, it may become trapped in local optima. To resolve this, we introduce a fallback strategy triggered when no informative viewpoints remain.  If the current domain is fully explored, a path to the global goal is generated by running the A$^{*}$ algorithm on the connectivity graph $\mathcal{G}_{R}$. Viewpoints are then selected along the resulting region path, and collision-free trajectories are computed. If a path between consecutive regions is blocked, a voxel-level A$^{*}$ path toe the next region's RP is used instead. This process continues until the global goal is reached.



\vspace{-1pt}

\subsection{Integrated Planning Framework}

For the integrated planning framework, if the current goal has been reached, the GP procedure is triggered.
The connectivity graph $\mathcal{G}_{C}$ is updated, and  region states are  refined based on frontier voxels. Solving the global TSP the provides a new region-level path.
The LP  activates after the robot completes the previous trajectory, constructing a target viewpoints graph $\mathcal{G}_{V}$ with local goal $x_{l,g}$. If no target viewpoints exist, the GP provides an alternative global path; otherwise, the LP generates the camera trajectory.


\vspace{-1pt}

\begin{figure}[t!]
        \centering
        \subfloat[]{
		\includegraphics[width=1.0\columnwidth]{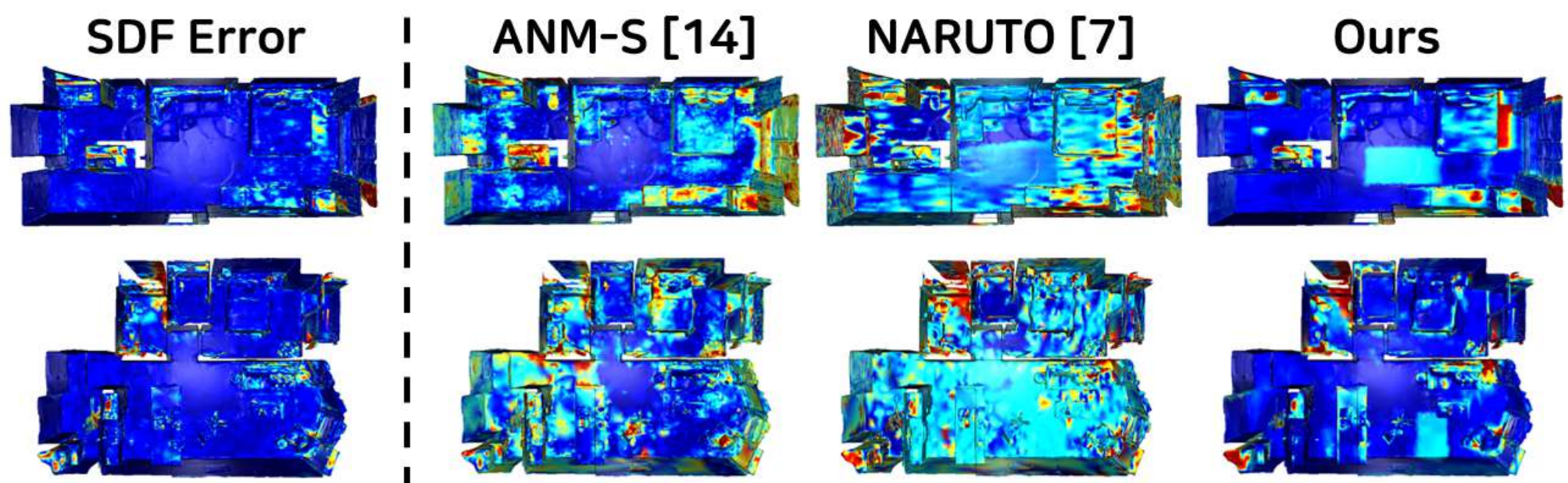}}
        \\
        \subfloat[]{
		\includegraphics[width=1.0\columnwidth]{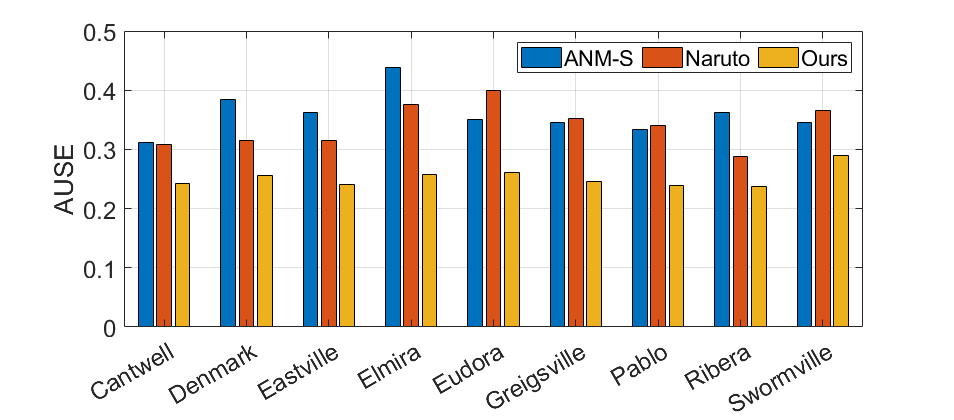}}
		\caption{(a) Uncertainty and SDF prediction error of the ground truth mesh on scenes from the Gibson dataset (Top: Pablo, Bottom: Swormville) for each algorithm during exploration. 
        Each value is normalized, with \textcolor{red}{red} (1) indicating a high value and \textcolor{blue}{blue} (0) indicating a low value.
        (b) AUSE plot for the UQ metrics evaluated on selected scenes from the Gibson dataset. 
        }
        \vspace{-4mm}
		\label{fig:uq_eval}
\end{figure}

\section{Experiments}
\vspace{-1pt}

\subsection{Experiment setup}
\vspace{-1pt}
\subsubsection{Implementation details}
We use a fixed parameter set across all experiments. The evidence scale $N_S$ is set to $e^{15}$, and $V_\rho$ is initialized to $-\log(N_S)$, yielding a fixed initial evidence value ($\approx\!1$) for unobserved voxels. $V_\rho$, $V_\tau$, and the voxel grid used for uncertainty accumulation all use a 0.1 m resolution. The region size for hierarchical planning is 1 m. The weight $k$ in the coverage planning cost function is 0.1. For target-viewpoints extraction, positions are sampled every 0.2 m in $x$-$y$ and 1 m in $z$, with 30 orientations per position, and the view selection threshold $\eta$ is 10.

\subsubsection{Dataset and Evaluation Environment} We evaluate our system using the Habitat simulator \cite{habitat19iccv} and photorealistic indoor scene datasets. 
We select five scenes from the MP3D dataset \cite{Matterport3D} and nine scenes from the Gibson dataset \cite{xiazamirhe2018gibsonenv}, which are the same scenes as \cite{yan2023active}. 
The planning loop is executed for 5000 time steps on MP3D and 1000 or 2000 time steps on Gibson, depending on scene size.\footnote{For the choice of time steps, we followed the evaluation in \cite{yan2023active}.} 
The system processes RGB-D images at the resolution of $680 \times 1200$ with the focal length of 600.
The voxel size for the 3D signed distance field is set to 0.1m.
All experiments are done with an Intel i7-10700 CPU and an NVIDIA RTX A5000 GPU.

\subsubsection{Evaluation Metrics}
The UQ module is evaluated using the Area Under the Sparsification Error (AUSE) \cite{ilg2018uncertainty}, computed as the area between error curves from two sparsification processes: one using the SDF values of ground-truth mesh vertices and the other using their uncertainty. A low AUSE indicates that the predicted uncertainty aligns more closely with the actual prediction error.
To assess the quality of the mesh, we follow the metrics used in previous works \cite{yan2023active, feng2024naruto},
completion (cm) and completion ratio (\%) with 5 cm threshold.
We apply mesh culling \cite{wang2023coslam} to remove unobserved regions and points that are outside the total scene.

\vspace{-1pt}

\subsubsection{Baseline Algorithms}
The UQ module is compared with those from other implicit neural representation (INR)-based active reconstruction methods \cite{feng2024naruto, kuang2024active}.
For assessing the overall system, we include frontier-based exploration \cite{yamauchi1997frontier} and another INR-based method \cite{yan2023active} as additional baselines. 

\begin{table}
	\centering
        \caption{Completion metrics for the Gibson and MP3D datasets. $\dagger$ indicates metrics from meshes converted from 3D Gaussians for a surface reconstruction comparison.}
	\label{tab:baselines}
    \resizebox{0.9\columnwidth}{!}{
	\begin{tabular}{lcccc}
		\toprule
		& \multicolumn{2}{c}{\emph{Gibson}} & \multicolumn{2}{c}{\emph{MP3D}} \\
		\cmidrule(lr){2-3} \cmidrule(lr){4-5}
		& \textbf{Comp. ↑} & \textbf{Comp. ↓} & \textbf{Comp. ↑} & \textbf{Comp. ↓}\\
		& (\%) & ($cm$) & (\%) & ($cm$) \\
        \toprule
		\textbf{FBE \cite{yamauchi1997frontier}} &  68.91 & 14.42 & 71.18  & 9.78 \\
        \cmidrule(lr){1-5}
		\textbf{ANM \cite{yan2023active}} & 80.45 & 7.44  & 73.15 & 9.11 \\
		\cmidrule(lr){1-5}
        \textbf{ANM-S \cite{kuang2024active}} & 92.10 & 2.83  & 89.74 & 4.14 \\
		\cmidrule(lr){1-5}
        \textbf{Naruto \cite{feng2024naruto}} & 90.31 & 4.31  & 90.18 & 3.00 \\
		\cmidrule(lr){1-5} \morecmidrules \cmidrule(lr){1-5}
        \textbf{FisherRF$^{\dagger}$} \cite{jiang2024fisherrf} & 82.24 & 5.06 & 80.26 & 6.55 \\
        \cmidrule(lr){1-5}
        \textbf{ActiveSplat{$^{\dagger}$}} \cite{li2025activesplat} & 84.71 & 5.41  & 76.71 & 5.26 \\
		\cmidrule(lr){1-5} \morecmidrules \cmidrule(lr){1-5}
        \textbf{Ours} & \textbf{93.49} & \textbf{2.60}  & \textbf{92.22} & \textbf{2.90} \\
        \bottomrule
	\end{tabular}
    }
    \vspace{-4mm}
\end{table}

\vspace{-2pt}

\subsection{Evaluation of Uncertainty Quantification}
For uncertainty quantification, we evaluated our epistemic UQ method against other INR-based active reconstruction methods, using both qualitative and quantitative analyses. 
For a fair comparison, all methods were trained using the same trajectories obtained from \cite{feng2024naruto}, and the AUSE was computed every 500 steps and averaged over the entire run.
As depicted in Fig. \ref{fig:uq_eval}a, our qualitative analysis visualizes the uncertainty metric and SDF error across two different scenes. 
Compared with the baselines, our uncertainty metric, which accounts for epistemic uncertainty, showed a strong correlation with SDF errors, indicating that regions with high uncertainty correspond to areas of lower accuracy. 
As shown in Fig. \ref{fig:uq_eval}b, our UQ outperforms the baselines across all selected scenes.

\begin{figure}[t!]
        \centering
		\includegraphics[width=0.85\columnwidth]{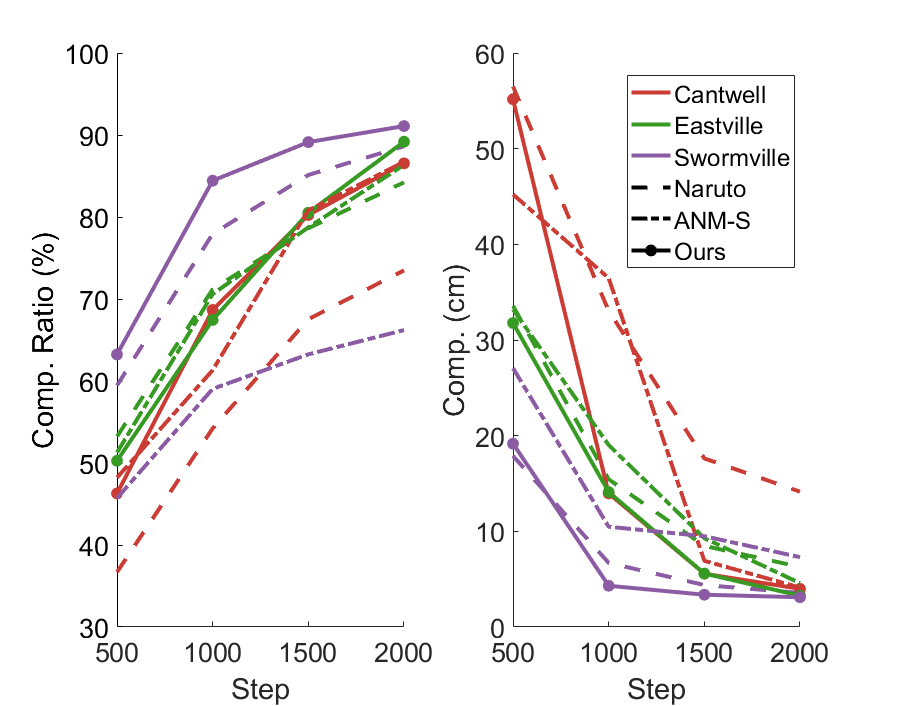}
		\caption{Comparison of completion ratio and completion metrics with respect to time for scenes of the Gibson dataset. 
        }
        \label{fig:comp_ratio}
		\vspace{-4mm}
\end{figure}

\begin{figure*}[t!]
  \centering
\includegraphics[width=1.95\columnwidth]{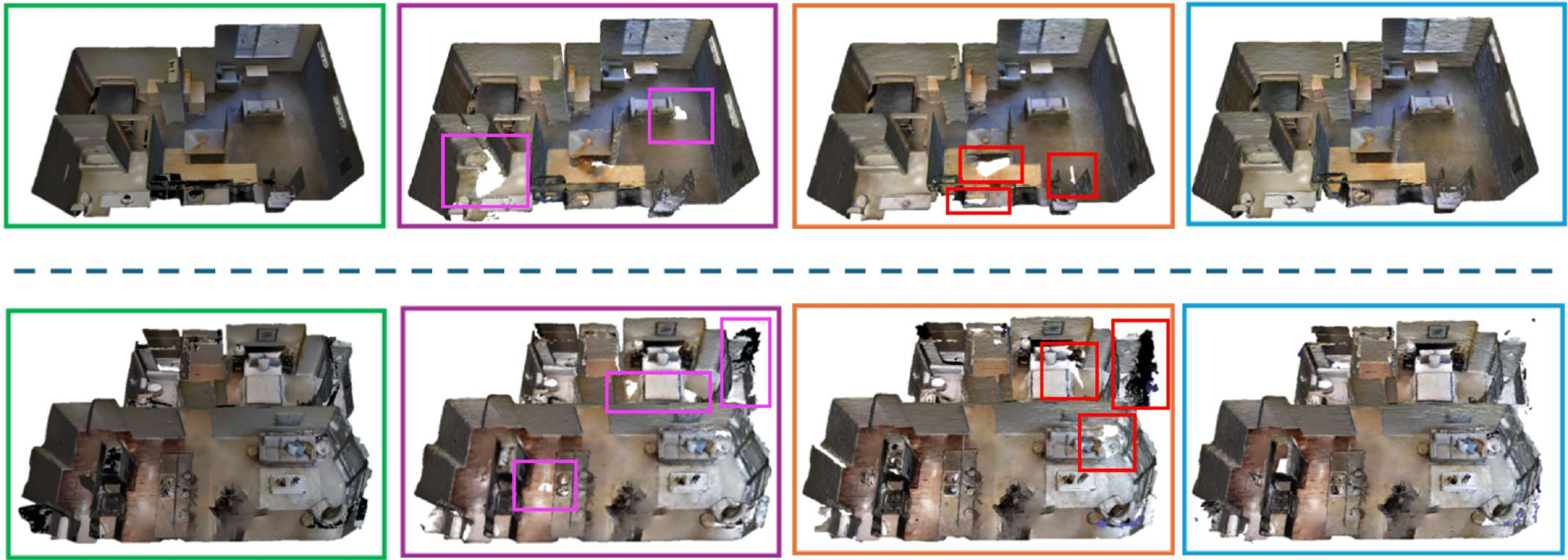}
  \caption{Reconstructed meshes from different active reconstruction methods (\textcolor{Green}{GT}, \textcolor{Purple}{ANM-S} \cite{kuang2024active}, \textcolor{Orange}{Naruto \cite{feng2024naruto}}, \textcolor{blue}{Ours}) for scenes from the Gibson dataset (Top: \textit{Ribera}, Bottom: \textit{Swormville}). The magenta and red boxes indicate regions where our method achieves more accurate reconstructions than \cite{kuang2024active} and \cite{feng2024naruto}, respectively, effectively avoiding artifacts and empty holes.}
  \vspace{-3mm}
  \label{fig:result_mesh}
\end{figure*}


\subsection{Evaluation of Active Scene Reconstruction} \label{sec:eval recon}
A brief overview of the evaluation on the completion of the active reconstruction methods is given in Table \ref{tab:baselines}. 
More detailed assessments of individual scenes can be found in the supplementary material. 
The result shows that the proposed method is superior in terms of completion metrics in the selected datasets. 
As can be observed in Fig. \ref{fig:result_mesh}, our method produces a more meticulously reconstructed mesh. 

Fig. \ref{fig:comp_ratio} compares our approach with the state-of-the-art INR-based active reconstruction system \cite{feng2024naruto} over different time steps of the planning loop on the Gibson dataset, focusing on complex scenes with more than six rooms. We applied the same mesh culling procedure used in our method to all baselines at each evaluation step.
Our method generally outperforms the baseline, except in the early stages of reconstruction. This is due to the hierarchical planning structure —- initially, the local planner dominates, prioritizing nearby information over global exploration. 
As reconstruction progresses, the global planner guides broader scene coverage while the local planner refines view selection, leading to superior completion metrics over time.

To validate the superiority of surface reconstruction over 3DGS-based methods, we report comparison with \cite{jiang2024fisherrf, li2025activesplat} in Table~\ref{tab:baselines}. To ensure fairness, we convert the 3D Gaussians into a mesh following \cite{huang20242d}, by rendering depth images at the training viewpoints, feeding them into TSDF fusion \cite{newcombe2011kinectfusion}, and extracting the scene mesh. Our method achieves higher completion metrics than 3DGS-based methods, indicating a more complete surface reconstruction.

Our system operates at 9.2 FPS. A single mapping, UQ over the entire scene, global planning, local planning, and escaping-local-minima step take 204, 2.28, 28.5, 16.1, and 158 ms, respectively, in the Gibson–Eudora scene.

\vspace{-1mm}

\subsection{Ablation Study} \label{sec:ablation}
Since the Gibson scenes include environments with varying scales and number of rooms, scene reconstruction becomes particularly challenging. Therefore, we conduct ablation studies on this dataset.

We evaluate the efficacy of the proposed UQ method as a plug-and-play enhancement that can be integrated with various active reconstruction planners requiring uncertainty quantification. 
We assess the active reconstruction performance of a planner adapted from \cite{feng2024naruto}, where our UQ method is employed for the goal search process. 
The average results on the Gibson dataset for the selected scenes are presented in Table \ref{tab:ablation_uq}. 
The improved reconstruction results across all metrics, compared to \cite{feng2024naruto}, demonstrate that the proposed UQ enhances the ability to identify informative goals for active reconstruction. 
This suggests that inferring epistemic uncertainty is crucial for informative path planning and highlights the versatility of the proposed UQ method.


\begin{table}[t!]
\caption{Ablation study on UQ in the Gibson dataset.
}
\centering
\label{table1}
\resizebox{0.7\columnwidth}{!}{%
\begin{tabular}{c|cccc}
\hline
& Comp. ↓ & Comp. Ratio ↑ \\ 
& (cm) &  (\%)
\\ \hline
\cite{feng2024naruto} & 4.31 & 90.31
\\ 
\cite{feng2024naruto} w/ Our UQ & 
2.69 & 
92.69
\\ 
Ours & \textbf{2.60} & \textbf{93.49} 
\\
\hline
\end{tabular}%
}
\label{tab:ablation_uq}
\vspace{-3mm}
\end{table}




\subsection{Real-world Experiments}
To demonstrate the efficacy of the proposed framework, we conducted hardware experiments using a ground robot (Turtlebot3 Waffle) equipped with a RGB-D camera (Realsense D455), with pose information provided by a motion capture system. To adapt the 6D planning framework for a ground vehicle, we sampled 3D target viewpoints. The proposed framework was tested in an indoor lab environment, and the resulting scene reconstruction is shown in Fig. \ref{fig:experiment}. Further details are provided in the supplementary video.

\begin{figure}[t!]
        \centering
		\includegraphics[width=0.99\columnwidth]{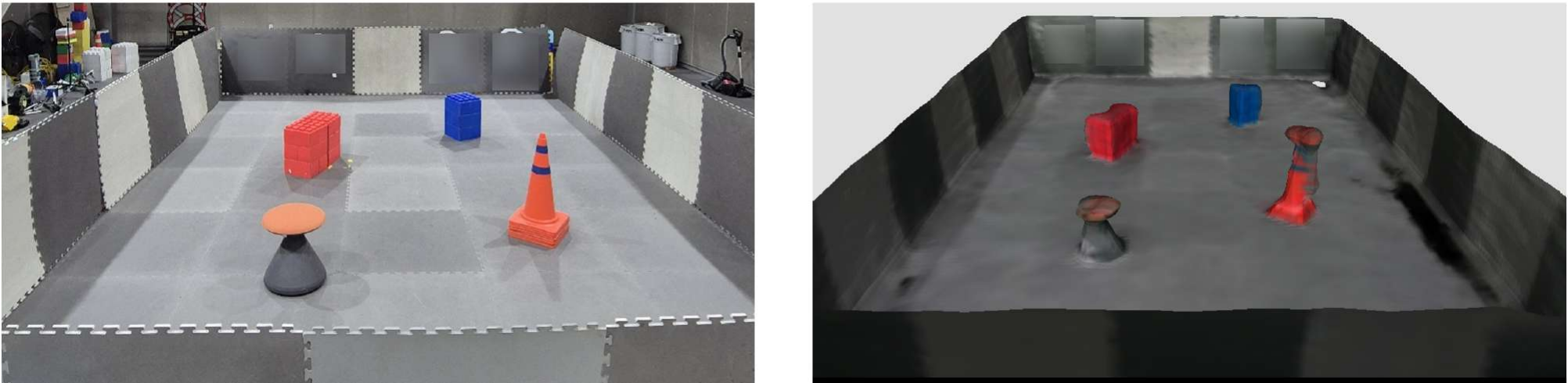}
		\caption{Result from the real-world experiment. The experimental scene (left), and the reconstructed mesh (right). 
        }
		\vspace{-6mm}
        \label{fig:experiment}
\end{figure}

\section{Conclusion and Discussion}

We presented an active 3D scene reconstruction method based on NeRF by introducing epistemic uncertainty through EDL. This enables real-time online planning with uncertainty estimates that align well with model error, effectively distinguishing epistemic from aleatoric uncertainty.
By combining a global coverage planner with a local, information-driven planner, our system scales effectively to large environments while leveraging epistemic uncertainty and adaptively generating camera trajectories that maximize information gain.
Our proposed method identifies informative trajectories that reduce human intervention in data collection, achieve more complete reconstructions than prior approaches, and yield high-quality meshes for downstream tasks.
Future work includes removing the assumption of prior localization for reconstruction in unseen environments~\cite{kim2021topology}, and incorporating object-level information~\cite{lee2024category} to support high-level tasks.


\bibliographystyle{./bibtex/IEEEtran}
\bibliography{./bibtex/IEEEabrv, ./bibtex/mybibfile} 

\addtolength{\textheight}{-12cm}   





\end{document}